\documentclass[journal]{IEEEtran}
\usepackage{url}
\usepackage{graphicx}
\usepackage{booktabs}
\usepackage{multirow}
\usepackage{array}
\usepackage{geometry}
\usepackage{amsthm}
\usepackage{amsmath,amssymb}
\usepackage{hyperref}
\ifCLASSINFOpdf
\else
\fi

\begin{document}
\title{TCMA: Text-Conditioned Multi-granularity Alignment for Drone Cross-Modal Text-Video Retrieval}

\author{
Zixu~Zhao, Yang~Zhan, Yunhao~Li, and Yan~Li
\thanks{Manuscript received , 2026. \textit{(Corresponding author: Yang Zhan)}}
\thanks{Zixu Zhao is with Carnegie Mellon University, Pittsburgh, PA 15213 USA (e-mail: clairez2@andrew.cmu.edu).}%
\thanks{Yang Zhan is with The Hong Kong Polytechnic University, Hong Kong, China (e-mail: zhanyang750@gmail.com).
% the Department of Logistics and Maritime Studies, 
}
\thanks{
Yunhao Li and Yan Li are with the Wuhan University, Wuhan 430079, China. 
% State Key Laboratory of Information Engineering in Surveying, Mapping and Remote Sensing, (e-mail: 2025106190002@whu.edu.cn, liyanWHU@whu.edu.cn)
}
}


\markboth{SUBMITTED to IEEE Transactions on Geoscience and Remote Sensing}%
{Shell \MakeLowercase{\textit{et al.}}: Bare Demo of IEEEtran.cls for IEEE Journals}



\maketitle

\begin{abstract}
Unmanned aerial vehicles (UAVs) have emerged as powerful platforms for real-time, high-resolution data collection, generating large quantities of aerial videos. Efficient text-video retrieval is crucial for applications including urban management, emergency response, security, and disaster relief. Research on UAV-based retrieval remains underexplored, as existing datasets suffer from coarse-grained and redundant captions. In this work, we construct a fine-grained Drone Video-Text Match Dataset (DVTMD), consisting of 2,864 videos and 14,320 semantically diverse captions. To foster this task, we propose the Text-Conditioned Multi-granularity Alignment (TCMA) framework, which integrates global video-sentence alignment, sentence-guided frame aggregation, and word-guided patch alignment. To further refine local alignment, we design a word and patch selection module that filters irrelevant content, as well as a text-adaptive dynamic temperature mechanism that adapts attention sharpness to text type. Extensive experiments on DVTMD and CapERA establish the comprehensive benchmark for drone text-video retrieval. Our TCMA achieves state-of-the-art results, with R@1 scores of 45.5\% for text-to-video retrieval and 42.8\% for video-to-text retrieval. 
\end{abstract}


% \fi
\IEEEpeerreviewmaketitle

\section{Introduction}
\IEEEPARstart{U}{nmanned} aerial vehicles (UAVs), commonly known as drones, have rapidly emerged as cost-effective and versatile platforms for high-resolution, real-time aerial data collection. 
With the global popularity of drones, massive aerial videos have been generated.
Therefore, efficient retrieval of relevant content from these large-scale UAV videos has become a critical research issue \cite{huang2024visual,zhan2026does,zhang2023hypersphere}. Drone text-video retrieval (DTVR) links natural language queries to video data. It is vital in domains such as urban management, emergency response, security surveillance, and disaster relief \cite{zhan2026mvpc}.

\begin{figure}[!t]
\centering
\includegraphics[width=0.9\linewidth]{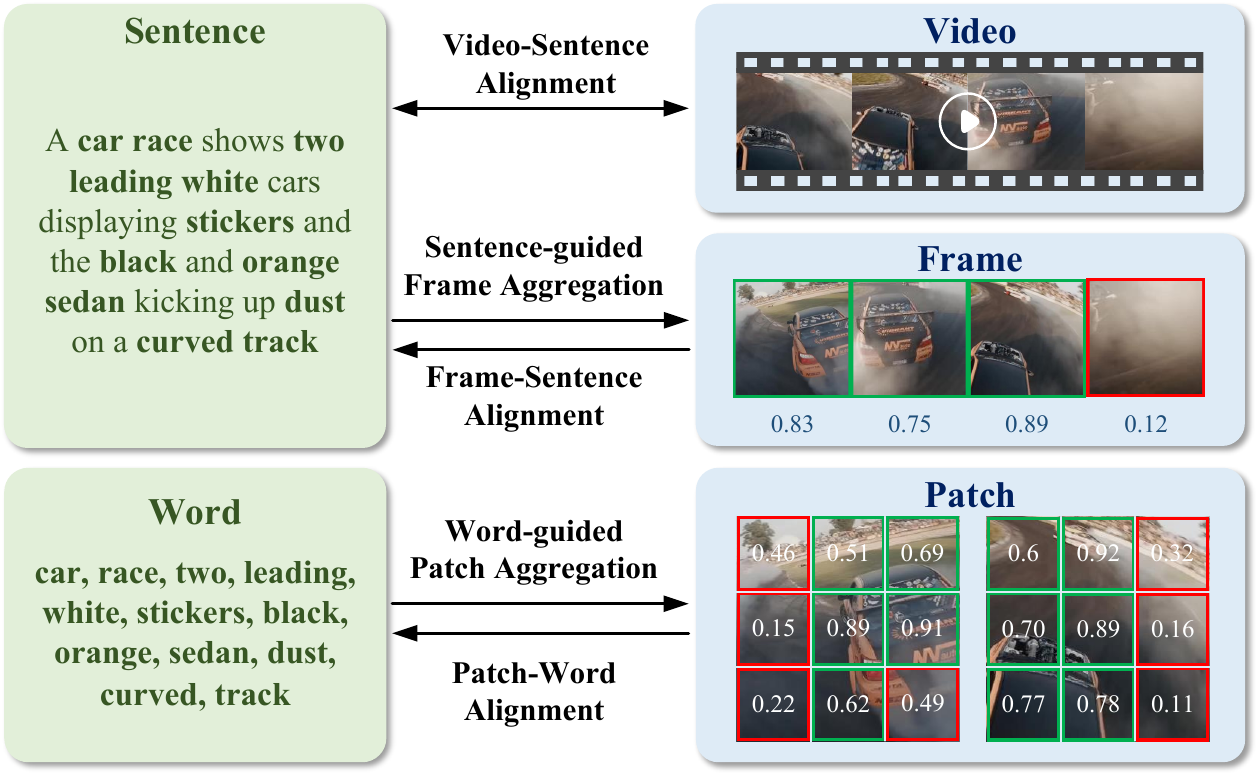}
\caption{Illustration of the characteristic in the drone text-video retrieval task.
Video-level: sentence and video encompass the overall semantics.
Frame-level: due to the rapid movement and wide field of view of the drone, there are many frames with low relevance to the query text that need to be filtered.
Patch-level: since video frames often contain extensive background regions and irrelevant objects, only a few patches are semantically relevant to the words in the query text.
}
\label{fig:intro}
\end{figure}

Although text-video retrieval has achieved substantial progress in natural video domains, such as MSR-VTT \cite{xu2016msr} and MSVD \cite{chen2011collecting}, the UAV video domain remains underexplored. A representative attempt to address this problem is CapERA \cite{bashmal2023capera}, which extends the ERA dataset \cite{mou2020era} with human-annotated captions. However, CapERA has two major limitations. First, its captions are coarse-grained. 
Videos within the same class often share highly similar descriptions that emphasize generic motion patterns while neglecting contextual factors such as environmental conditions and background objects. Second, the captions lack semantic diversity. Many are generated by paraphrasing or back-translation \cite{bashmal2023capera} of a single source description, leading to severe redundancy among multiple captions for the same video. This lack of granularity and diversity weakens text-video correspondence and severely limits retrieval models’ ability to discriminate among visually similar aerial videos. Consequently, existing aerial datasets remain inadequate for supporting accurate and robust UAV text-video retrieval.

To overcome these limitations, we present the Drone Video-Text Match Dataset (DVTMD) with fine-grained and semantically diverse captions. DVTMD contains 2,864 videos and 14,320 captions, with five unique descriptions assigned to each video. Unlike prior datasets that rely on generic motion-oriented labels and captions, our annotations comprehensively cover five complementary and fine-grained dimensions of each video: (\textit{i}) human figures and their actions, (\textit{ii}) objects and equipment present in the scene, (\textit{iii}) background settings and structural elements, (\textit{iv}) weather and environmental conditions, and (\textit{v}) overall visual tone and stylistic attributes.
By systematically integrating these dimensions, DVTMD provides richer semantic coverage and reduces redundancy across captions. This fine-grained annotation strategy strengthens text-video correspondence and facilitates more precise discrimination among visually similar aerial videos.

Recent advances in large-scale vision-language pretraining, exemplified by CLIP \cite{radford2021learning}, have demonstrated strong image-text alignment capabilities. Building upon this foundation, CLIP4Clip \cite{luo2022clip4clip} adapts CLIP to the video domain via temporal pooling, while subsequent works \cite{gorti2022x, liu2022ts2, fang2023uatvr, shen2024tempme} refine alignment by modeling visual-text interactions. However, their effectiveness does not fully transfer to drone text-video retrieval. As shown in Figure \ref{fig:intro}, UAV videos, captured from aerial viewpoints, cover wide scenes with multiple moving targets and complex backgrounds. Compared with ground-view videos, they involve larger spatial scales, diverse object sizes, and dynamic environments. CLIP-based methods emphasize global cues such as scene layout or dominant motion, but this coarse alignment often fails, as brief captions typically describe only subclips or subregions. 
To overcome this limitation, we introduce fine-grained alignment by establishing correspondences between video patches and word tokens. This approach is capable of capturing localized details, such as individual objects, vehicles, or human actions. However, it often overlooks higher-level contextual information, including the broader environment, spatial layout, and collective activity within the scene. For UAV scenarios, both global semantics and local details are indispensable. 

Motivated by these challenges, we propose a Text-Conditioned Multi-granularity Alignment (TCMA) framework. At the global level, coarse video-sentence alignment captures overall semantics. At the temporal level, sentence-guided frame aggregation emphasizes frames most relevant to the description. At the local level, word-guided patch alignment refines correspondence at the object and region scale. This hierarchical design jointly leverages global context and fine-grained detail, thereby achieving more accurate and robust drone text-video retrieval.

Within the proposed TCMA framework, we further refine local-level alignment through a Word and Patch Selection Module. Captions for UAV videos typically focus on specific objects or actions, while video frames often contain extensive background regions and irrelevant elements. As a result, only a limited portion of the visual content is semantically relevant. To address this imbalance, we select the most informative patches from each frame and the most salient words from each sentence. This targeted selection suppresses background noise and redundant textual tokens, such as common stopwords, thereby preventing attention dilution and reinforcing meaningful word-patch interactions.

Finally, we introduce a Text-Adaptive Dynamic Temperature Mechanism to regulate attention sharpness. Conventional approaches apply a static temperature to control attention distribution, yet different textual descriptions demand varying levels of focus. Action-oriented descriptions require sharper attention to highlight a few critical frames, whereas scene-oriented descriptions demand smoother attention to capture broader contextual cues. To address this, our module dynamically predicts the optimal temperature conditioned on each input text, enabling adaptive attention that flexibly adjusts focus according to description type, thereby enhancing retrieval robustness.

Our main contributions of the work are as follows:
\begin{itemize}
    \item We construct DVTMD, a drone video-text match dataset comprising 2,864 videos and 14,320 captions, which provides fine-grained and semantically diverse annotations across different dimensions.
    \item We propose a text-conditioned multi-granularity alignment framework that combines global video-sentence alignment, sentence-guided frame aggregation, and word-guided patch alignment to balance global context with fine-grained detail. 
    \item We design a word and patch selection module to extract the most informative visual patches and textual tokens. Meanwhile, we introduce a text-adaptive dynamic temperature mechanism to adjust attention sharpness according to the query text. 
    \item We conduct comprehensive evaluations on DVTMD and CapERA, and establish the standard benchmark for the drone text-video retrieval task.  Extensive experiments demonstrate that our method outperforms existing state-of-the-art approaches.
\end{itemize}

\section{Related Work}
In this section, we review prior work on drone video–text retrieval, covering aerial cross-modal methods in Section \ref{sec:aerial} and general text-video retrieval methods in Section \ref{sec:textvideo}.

\subsection{Cross-Modal Retrieval in Aerial Scenarios}
\label{sec:aerial}
Remote Sensing Image–Text Retrieval (RSITR) was the first step toward cross-modal understanding in aerial scenarios. It enables the retrieval of remote sensing data across visual and textual modalities. Early approaches relied on CNN backbones and recurrent networks trained from scratch \cite{abdullah2020textrs, yuan2022exploring, zhang2023hypersphere, pan2023prior, ji2023knowledge}. Abdullah et al. \cite{abdullah2020textrs} pioneered RSITR using an average fusion strategy, while Yuan et al. \cite{yuan2022exploring} introduced visual self-attention and released the RSITMD benchmark. Subsequent studies improved cross-modal alignment through knowledge-aided learning \cite{ji2023knowledge}, key-entity attention \cite{zhang2023hypersphere}, and language-cycle attention \cite{pan2023prior}. With the advent of foundation models (FMs) such as CLIP \cite{radford2021learning} and BLIP \cite{li2022blip}, research has shifted toward leveraging pretrained vision–language representations for RSITR. Recent efforts in parameter-efficient fine-tuning \cite{yuan2023parameter} and large-scale benchmarking \cite{ZHAN202564} further demonstrated the promise of FMs in this domain. Extending beyond satellite imagery, VCSR \cite{huang2024visual} introduced UAV Image–Text Retrieval by incorporating context-region learning and consistency-based semantic alignment. Recent work by Bashmal et al \cite{bashmal2023capera,bashmal2024capera+} proposed a dataset capERA and a text-to-event retrieval framework for aerial videos by combining a vision transformer (ViT)–based video branch with a BERT-based text branch, optimized via a bidirectional contrastive loss. 
% Nevertheless, drone video–text retrieval remains largely underexplored.

\subsection{Text-Video Retrieval}
\label{sec:textvideo}
Text-Video retrieval aims to find the most relevant video given a natural language query, or vice versa, retrieving text descriptions corresponding to a video. This task requires modeling both cross-modal correspondence and temporal dynamics in videos. Existing methods can be broadly categorized into single-stream, dual-stream, and CLIP-based approaches.

Single-stream method like ActionBERT \cite{he2021actionbert} jointly embed video-text pairs within a unified network, enabling early cross-modal fusion. While effective at capturing fine-grained interactions, these approaches often have high computational costs and limited adaptability to downstream tasks. In contrast, dual-stream methods employ separate encoders for video and text, followed by contrastive learning in a joint embedding space. This design improves cross-modal representation learning and training efficiency. 
Representative advances include MIL-NCE \cite{miech2020end}, which learns video–text representations from narrated videos without human annotation by introducing a multiple instance learning noise contrastive loss; ClipBERT \cite{lei2021less}, which reduces training cost via sparse clip sampling in end-to-end modeling; and Frozen in Time \cite{bain2021frozen}, which unifies image–text and video–text pretraining with a space–time transformer encoder.

More recently, CLIP-based approaches leverage the pre-trained language–vision model CLIP, which is trained on large-scale text–image data. Methods such as CLIP4Clip \cite{luo2022clip4clip} adopt CLIP as a backbone for video–text retrieval by directly mean-pooling frame-level features to obtain a compact video representation. Beyond global pooling, X-Pool \cite{gorti2022x} introduces cross-modal attention to allow texts to attend selectively to their most semantically relevant video frames, thereby avoiding noise from irrelevant visual content. Beyond discriminative matching, generative approaches have been explored. DiffusionRet \cite{jin2023diffusionret} extends the CLIP paradigm by modeling the joint distribution of video and text, combining generation and discrimination objectives. Cap4Video \cite{wu2023cap4video} leverages auxiliary captions generated from videos to enrich cross-modal alignment, incorporating captions as additional input data. To enhance scalability, parameter-efficient adaptation techniques have emerged. MV-Adapter \cite{jin2024mv} inserts lightweight adapters into CLIP layers to transfer knowledge with minimal trainable parameters, while VoP \cite{huang2024visual} introduces cooperative prompt tuning for both text and video, achieving strong performance with less than 0.1\% trainable parameters. TeachCLIP \cite{tianteachclip2024} distills fine-grained cross-modal reasoning from heavy teacher models into a CLIP4Clip-based student, maintaining efficiency while achieving near state-of-the-art accuracy. Fine-grained spatial–temporal modeling has also gained traction. TS2-Net \cite{liu2022ts2} dynamically adjusts token sequences to select informative tokens in both temporal and spatial dimensions. MCQ \cite{ge2022bridging} introduces a pretext task formulated as multiple choice questions on erased details of nouns and verbs to explicitly associate intermediate video and text tokens. UCofia \cite{wang2023unified} and EERCF \cite{tian2024towards} adopt coarse-to-fine strategies to refine video–text matching from global to local levels.

\section{DVTMD Dataset}
In this section, we introduce the DVTMD dataset. Section \ref{sec:construction} details the dataset construction process.
% , including frame sampling, frame-level description, video-level summarization, and final dataset formation. 
Section \ref{sec:analysis} presents a comprehensive analysis of DVTMD.
% , covering statistics, caption distribution, lexical diversity, and comparison with existing datasets.

\begin{figure*}[!t]
\centering
\includegraphics[width=0.9\linewidth]{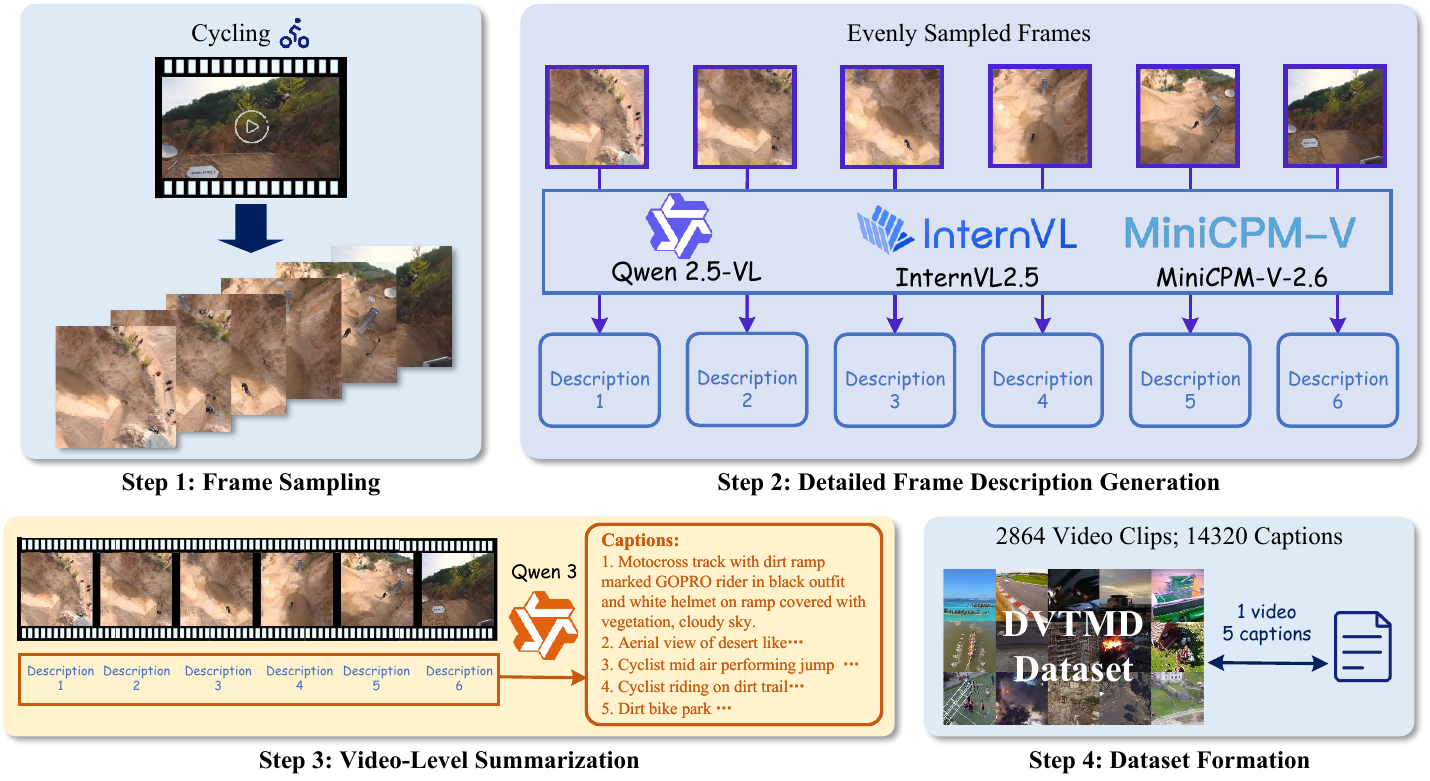}
\caption{The construction pipeline of our DVTMD benchmark: Step 1) frame sampling, Step 2) detailed frame description generation, Step 3) video-level summarization, and Step 4) dataset formation.}
\label{fig:dataset}
\end{figure*}

\subsection{Dataset Construction}
\label{sec:construction}
The construction pipeline of the newly proposed DVTMD is shown in Figure \ref{fig:dataset}. We choose the ERA dataset \cite{mou2020era} as our aerial video source. ERA comprises 2,864 video clips, each annotated with one of 25 event categories. Each clip lasts 5 seconds, recorded at 24 frames per second with a spatial resolution of 640$\times$640 pixels. To adapt this resource for multimodal retrieval tasks, we perform the following steps.

\textbf{Step 1: Frame Sampling.}  
To reduce redundancy while preserving temporal information, six evenly spaced frames are uniformly extracted from each video. Formally, given a video $V \in \mathbb{R}^{T \times H \times W \times 3}$, frames are sampled at $F_1 = V\left[\tfrac{T}{6}\right], \; F_2 = V\left[\tfrac{2T}{6}\right], \dots, \; F_6 = V[T]$.
This ensures a balanced frame set $V' \in \mathbb{R}^{6 \times H \times W \times 3}$ well-distributed across the entire video duration for subsequent annotation.

\textbf{Step 2: Detailed Frame Description Generation.}  
Each sampled frame is annotated using state-of-the-art vision-language models (VLMs). After comparative evaluation of Qwen2.5-VL-7B \cite{bai2025qwen25vltechnicalreport}, MiniCPM-V-2.6 \cite{yao2024minicpm}, and InternVL2.5-8B \cite{chen2025expanding}, we adopt Qwen2.5-VL-7B due to its higher consistency and accuracy. Our prompt design is driven by the need for objective, literal, and fine-grained descriptions that can distinguish subtle differences across frames within the same or similar video genres. Specifically, the prompts instruct the model to systematically describe human figures, objects, backgrounds, color tones, and spatial arrangements without inferring unseen events or emotional context. 
% The detailed prompt template is illustrated in Figure~\ref{fig:prompt1}.

\textbf{Step 3: Video-Level Summarization.}  
To condense frame-level descriptions into concise video-level captions, we employ Qwen3 \cite{yang2025qwen3} due to its strong performance in structured summarization tasks. The summarization prompt is carefully crafted to guide the model towards factual, dense, and temporally-aware descriptions covering subjects, actions, backgrounds, and distinctive details, while excluding imaginative or rhetorical content. Each video is summarized into five captions of approximately 20 words each. 
% The summarization prompt template is illustrated in Figure~\ref{fig:prompt2}.

\textbf{Step 4: Dataset Formation.}  
Through this process, we establish a one-to-many correspondence between videos and their associated captions. Each of the 2,864 ERA videos is paired with five structured captions, producing 14,320 captions in total. The resulting DVTMD dataset serves as the basis for subsequent retrieval and analysis tasks in our study.

% \begin{figure*}[!t]
% \centering
% \includegraphics[width=0.95\linewidth]{Figure_3.pdf}
% \caption{The designed prompt template for generating detailed and objective frame-level descriptions.}
% \label{fig:prompt1}
% \end{figure*}

% \begin{figure*}[!t]
% \centering
% \includegraphics[width=0.95\linewidth]{Figure_4.pdf}
% \caption{The designed prompt template for generating summarized captions.}
% \label{fig:prompt2}
% \end{figure*}

\subsection{Dataset Analysis}
\label{sec:analysis}
% The key statistics of DVTMD are summarized in Table~\ref{tab:dataset}. 
Our DVTMD contains 2,864 videos, with 1,473 videos designated for training and 1,391 videos for testing. Each video is labeled with one of 25 event categories. For each video, we collect exactly 5 captions, resulting in a total of 14,320 captions across the entire dataset, of which 7,365 belong to the training set and 6,955 to the test set. The captions are constructed to be dense and descriptive, with an average length of 21.65 words. The vocabulary size of DVTMD amounts to 7,471 unique words, reflecting the diversity of expressions and scene compositions captured by UAVs. To more comprehensively understand the characteristics of the constructed DVTMD, we conduct a graphical analysis of the captions from the perspectives of caption length, category distribution, and lexical coverage.

% \begin{table}[!t]
% \caption{Dataset Statistics}
% \label{tab:dataset}
% \centering
% \small
% \setlength{\tabcolsep}{5pt}
% \begin{tabular}{lc}
% \toprule
% \textbf{Statistic} & \textbf{Value} \\
% \midrule
% Number of Videos & 2,864 \\
% Number of Categories & 25 \\
% Number of Captions & 14,320 \\
% Captions per Video & 5 \\
% Vocabulary Size & 7,471 \\
% Average Caption Length (words) & 21.65 \\
% \bottomrule
% \end{tabular}
% \end{table}

\begin{figure*}[!t]
\centering
\includegraphics[width=0.9\linewidth]{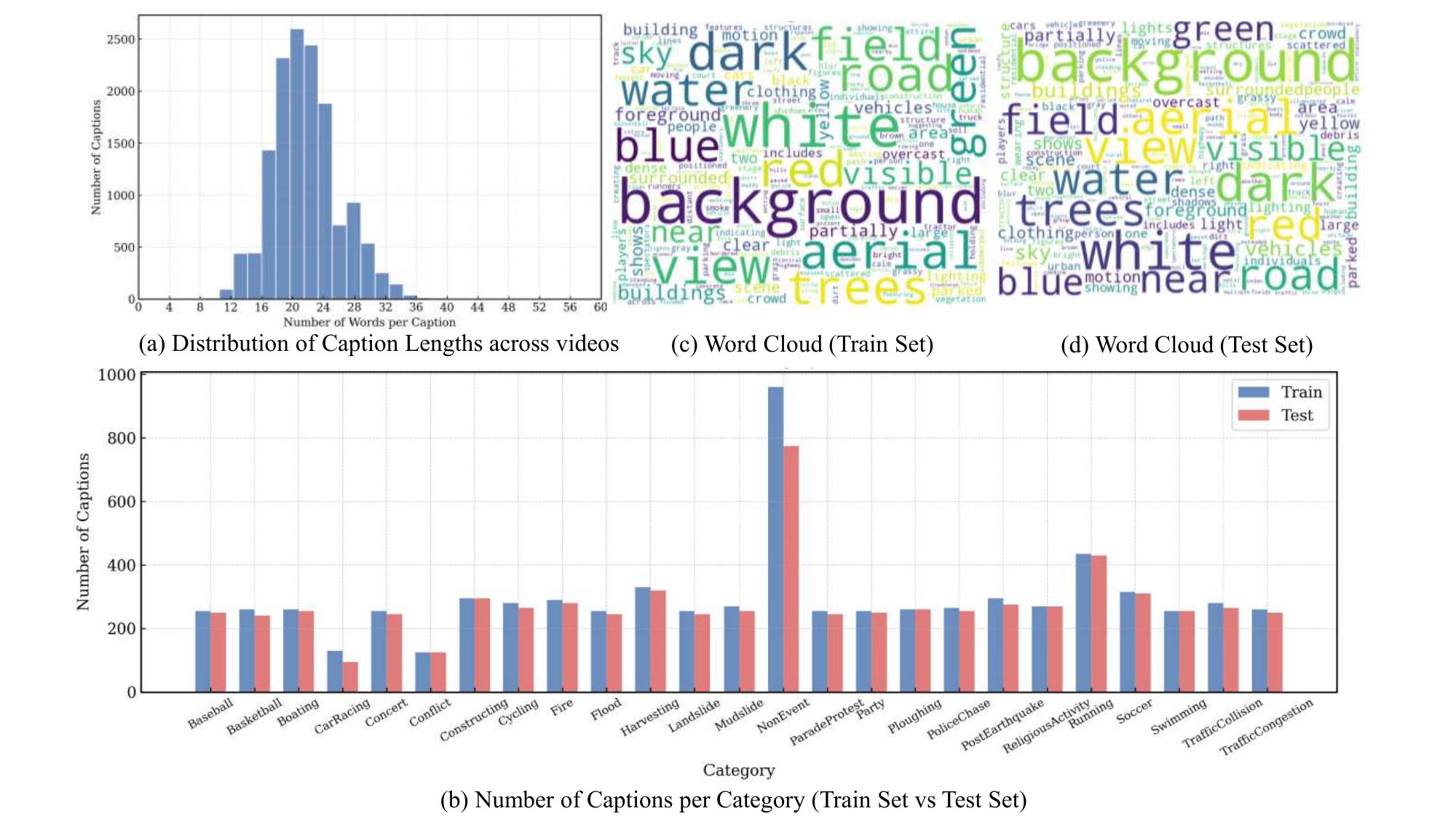}
\caption{Graphical analysis of the captions. (a) Caption length distribution across videos, with most captions falling between 15 and 25 words, peaking around 18–22 words. (b) Number of captions per category in the training and testing splits. The "NonEvent" category exhibits a minor discrepancy between the splits while other categories maintain a balanced distribution. (c),(d) Word clouds of the train set and test set captions displaying various elements, including humans, actions, objects, and backgrounds.}
\label{fig:comb}
\end{figure*}

\begin{figure*}[!t]
\centering
\includegraphics[width=0.9\linewidth]{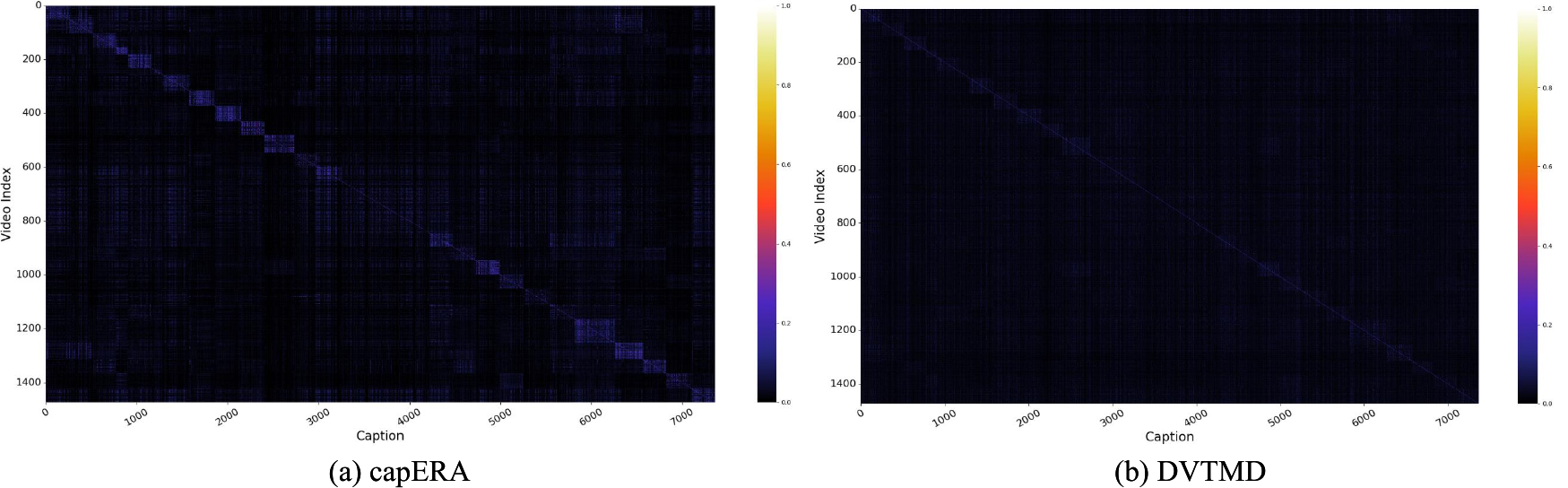}
\caption{Similarity heatmaps of video-caption pairs based on weighted BLEU and METEOR scores. (a) capERA exhibits extensive off-diagonal similarity, indicating that captions within the same category tend to be overly similar. (b) Our dataset shows a clearer diagonal structure, suggesting finer granularity and better distinction between captions of videos.}
\label{fig:heatmap}
\end{figure*}

Figure~\ref{fig:comb} (a) illustrates the caption length distribution. Most captions fall between 15 and 25 words, peaking at 18–22 words. This reflects the design goal of balancing clarity and descriptive richness. The slight skew toward shorter captions indicates a preference for concise, information-dense descriptions without excessive verbosity.Figure~\ref{fig:comb} (b) presents the number of captions per category across training and testing splits. Despite minor variations in the "NonEvent" category, which contains more captions in the test set than in the training set, DVTMD maintains a balanced distribution over 25 categories. Figure~\ref{fig:comb} (c) and Figure~\ref{fig:comb} (d) illustrate word clouds of train set and test set, respectively, generated from the captions after removing stop words unrelated to video content. These visualizations demonstrate the coverage of various elements, including humans, actions, objects, backgrounds, and environmental conditions.

\begin{figure*}[!t]
\centering
\includegraphics[width=0.9\linewidth]{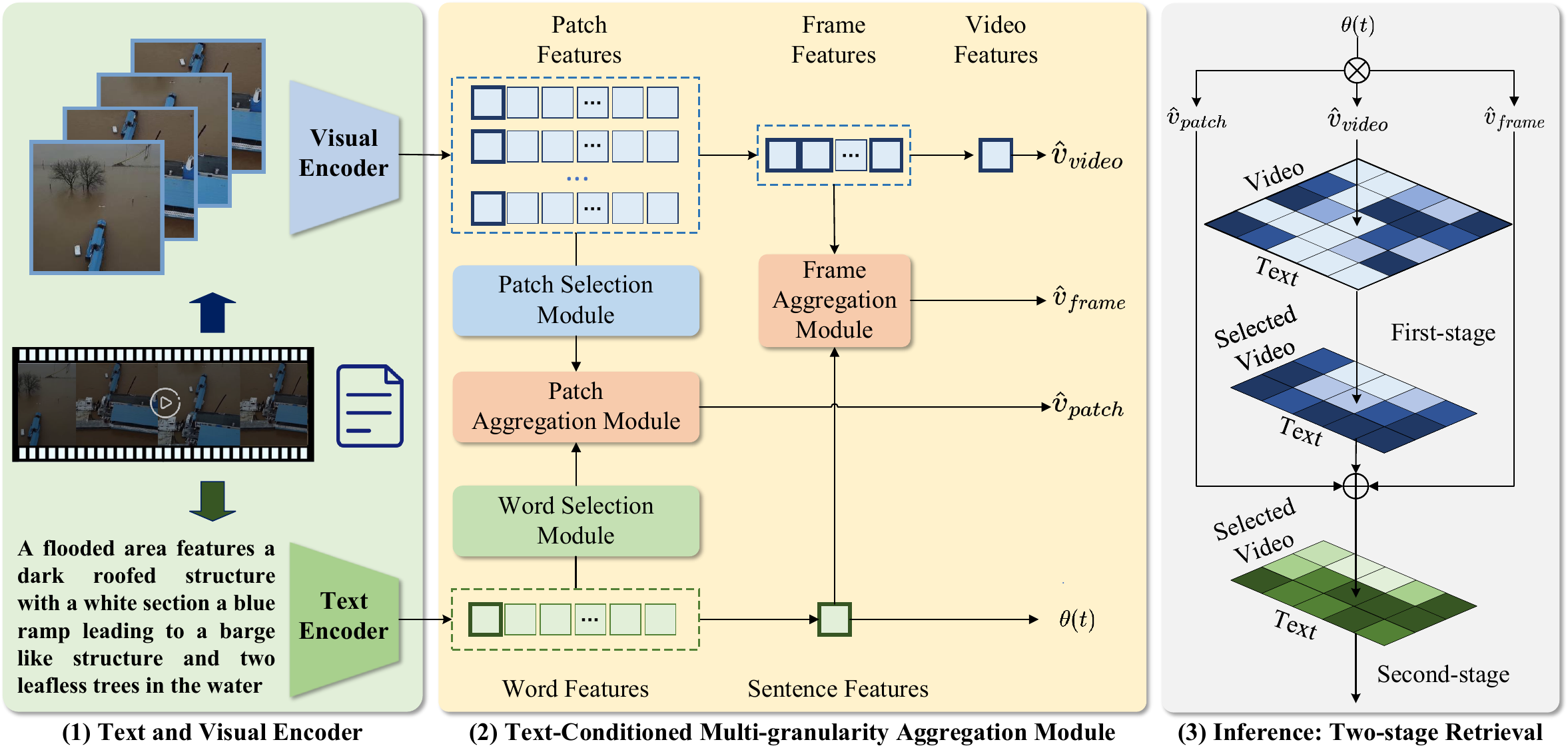}
\caption{
The overview of our proposed TCMA. It consists of three components: 
(1) Text and Visual Encoder to extract word features, sentence features, patch features, and frame features; 
(2) The Text-Conditioned Multi-granularity Aggregation Module, which progressively aligns video with text at three levels of granularity: text-agnostic video pooling ($\hat{v}_{video}$), sentence-guided frame aggregation ($\hat{v}_{frame}$), and word-guided patch aggregation ($\hat{v}_{patch}$);
(3) Two-stage Retrieval, which first performs coarse retrieval with global features and then refines alignment with fine-grained features for robust video-text matching during inference.
}
\label{fig:overview}
\end{figure*}

Finally, to assess the granularity of captions relative to the capERA dataset, we compute the weighted average BLEU and METEOR scores between video-caption pairs and visualize the results in Figure~\ref{fig:heatmap}. The goal is to ensure the captions are distinct and sufficiently detailed. Ideally, the similarity matrix would form a diagonal line from the top-left to the bottom-right, indicating that each caption corresponds only to its respective video. However, in Figure~\ref{fig:heatmap} (a), the similarity matrix for capERA shows significant off-diagonal lines and color blocks, suggesting that captions within the same category are overly similar. This observation underscores that our dataset, in contrast to capERA, has a clearer diagonal structure that better distinguishes videos with subtle variations within the same genre, offering a more nuanced representation of intra-category diversity.

\section{Methodology}
In this section, we introduce our proposed framework.
Section \ref{sec:pretrain} explains preliminary knowledge on how we adapt the CLIP model to the text-video retrieval task. 
Section \ref{sec:tcma} introduces our TCMA framework, including hierarchical alignment at video, frame, and patch levels. 
Section \ref{sec:loss} details the hierarchical alignment loss that integrates contrastive learning with pearson correlation regularization across multiple granularities.

\subsection{Expanding Language-Image Pre-training Model}
\label{sec:pretrain}
We adopt CLIP \cite{radford2021learning} Given a video set $V=\{v_1, \dots, v_N\}$ and a caption set $T=\{t_1, \dots, t_N\}$, the objective is to learn a similarity function $s(v_i, t_j)$ that produces high scores for semantically aligned pairs and low scores for mismatched ones.  
For each video $v_i$, a sequence of $T$ frames is uniformly sampled. Each frame $f_i^t$ is encoded by the pre-trained CLIP \cite{radford2021learning} encoder to obtain frame-level embeddings:
\begin{equation}
F_i = \{\phi(f_i^1), \phi(f_i^2), \ldots, \phi(f_i^T)\}
\end{equation} 
where $\phi(f_i^t) \in \mathbb{R}^D$.
The video-level representation is then obtained via mean pooling:
\begin{equation}
\hat{v}_i = \frac{1}{T}\sum_{t=1}^{T} \phi(f_i^t).
\end{equation}

For text input $t_j$, the Transformer-based encoder with 12 layers, width 512, and 8 attention heads is applied, and the output of the [EOS] token at the final layer is used as the sentence embedding:
$\theta(t_j) \in \mathbb{R}^D$.
The similarity between video and text is then calculated by cosine similarity:
\begin{equation}
s(v_i, t_j) = \frac{\theta(t_j)^\top \hat{v}_i}{\|\theta(t_j)\|\|\hat{v}_i\|}.
\end{equation}

\begin{figure*}[!t]
\centering
\includegraphics[width=0.9\linewidth]{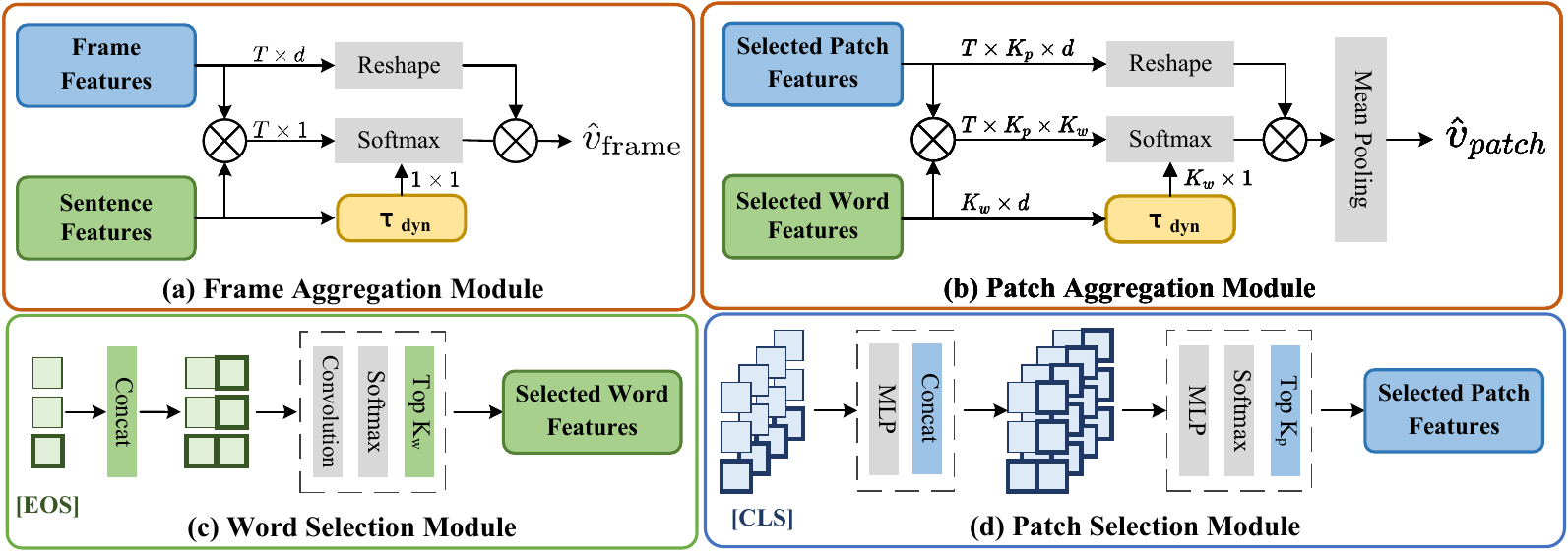}
\caption{(a) Frame Aggregation Module, which carries out sentence-guided frame aggregation to form $\hat{v}_{frame}$.(b) Patch Aggregation Module, which carries out word-guided patch aggregation to form $\hat{v}_{patch}$. (c) Word Selection Module, which highlights informative words. (d) Patch Selection Module, which filters out redundant background patches.}
\label{fig:overview2}
\end{figure*}

\subsection{Text-Conditioned Multi-granularity Aggregation}
\label{sec:tcma}

To effectively align video and text at different levels of granularity, we design a hierarchical aggregation strategy shown in Figure \ref{fig:overview} that progressively enhances video representations with textual guidance. Specifically, we introduce three complementary modules: (1) \textit{text-agnostic video pooling} to capture holistic video semantics, (2) \textit{sentence-guided frame aggregation} to adaptively weight frames according to sentence relevance, and (3) \textit{word-guided patch aggregation} to model fine-grained correspondence between salient words and patches. 

\subsubsection{Text-Agnostic Video Pooling}
We first obtain a video-level representation that is independent of text. Specifically, we apply mean pooling over all frame features to produce a global video embedding $\hat{v}_{\text{video}}$:
\[
\hat{v}_{video} = \frac{1}{T}\sum_{t=1}^{T} \phi(f^t),
\]
where $\hat{v}_{\text{video}} \in \mathbb{R}^D$ serves as the aggregated representation of video $v$.

\subsubsection{Sentence-Guided Frame Aggregation}
To model the cross-modal correspondence between a sentence and the video frames, we propose a dynamic sentence-guided Frame Aggregation Module, inspired by EERCF \cite{tian2024towards}. 
As shown in Figure \ref{fig:overview2} (a), this module adaptively weights each video frame based on its semantic relevance to the input sentence, producing a sentence-guided video representation $\hat{v}^{\text{frame}}$.

Given a sentence feature vector $\theta(t) \in \mathbb{R}^D$ and a sequence of $T$ frame features $F = \{\phi(f^1), \phi(f^2), \dots, \phi(f^T)\}$ with $\phi(f^i) \in \mathbb{R}^D$, the frame aggregation module computes sentence-to-frame similarity through dot products:
\begin{equation}
s_i = \phi(f^i)^\top \theta(t), \quad i = 1, 2, \dots, T.
\end{equation}

The similarity scores are normalized into attention weights using a softmax function with temperature $\tau$:
\begin{equation}
a_i = \frac{\exp(s_i / \tau)}{\sum_{j=1}^{T} \exp(s_j / \tau)}.
\end{equation}

The sentence-guided video embedding is obtained by weighted aggregation:
\begin{equation}
\hat{v}_{\mathrm{frame}} = \sum_{i=1}^{T} a_i \, \phi(f^i).
\end{equation}

Here, a smaller $\tau$ emphasizes a few highly relevant frames, while a larger $\tau$ distributes attention more evenly across frames. Different sentences may require varying attention sharpness: action-oriented sentences benefit from smaller $\tau$ for concentrated attention, whereas scene-oriented sentences benefit from larger $\tau$ for smoother aggregation. To address this, we introduce a lightweight Text-Adaptive Dynamic Temperature Mechanism. The sentence feature $\theta(t)$ is input to a small network to predict the optimal temperature:
\begin{equation}
\tau_{\text{dyn}}(t) = \mathrm{Softplus}(W_\tau \theta(t) + b_\tau) + \epsilon,
\end{equation}
where $W_\tau \in \mathbb{R}^{1 \times D}$ and $b_\tau$ are learnable parameters, Softplus ensures positivity, and $\epsilon=0.001$ prevents numerical instability. The attention weights are then recomputed using $\tau_{\text{dyn}}(t)$:
\begin{equation}
\hat{a}_i = \frac{\exp(s_i / \tau_{\text{dyn}}(\theta(t)))}{\sum_{j=1}^{T} \exp(s_j / \tau_{\text{dyn}}(\theta(t)))}.
\end{equation}

Finally, the compact formulation of frame-level representation is expressed as:
\begin{equation}
\hat{v}_{\mathrm{frame}}
= \sum_{i=1}^T 
\mathrm{Softmax}\!\Bigg(
\frac{\phi(f^i)^\top\theta(t)}{\tau_{\mathrm{dyn}}(t)}
\Bigg) \; \phi(f^i).
\end{equation}

This design enables frame aggregation to adaptively focus on the most relevant frames for each sentence and produce the most appropriate frame-level representation, improving sentence-to-video alignment without introducing heavy additional parameters.

\subsubsection{Word-Guided Patch Aggregation}
High-level frame features often fail to capture subtle cross-modal correspondence. To achieve fine-grained alignment, we leverage patch-level visual tokens and word-level textual tokens. However, directly aligning all patches with all words is inefficient due to the redundancy of background patches. To address this, we design a Word and Patch Selection Module, as shown in Figure \ref{fig:overview2} (c) and Figure \ref{fig:overview2} (d). They select the most salient patches and words for alignment.

\textbf{Word Selection Module.}
Given the word embeddings $T = \{\theta(w_1), \dots, \theta(w_L)\}$ with $\theta(w_l)\in\mathbb{R}^D$, 
we first augment each patch with the sentence feature $\theta(t)$ then compute a saliency score for each word:
\begin{equation}
u_l^{\text{word}} = G_w(\mathrm{Concat}(\theta(w_l),\theta(t))), \quad l = 1,\dots,L,
\end{equation}
where $G_w$ is a convolutional layer.  
Based on $u_l^{\text{word}}$, we select the top-$K_w$ informative words:
\begin{equation}
\hat{T} = \mathrm{TopK}(T, \{u_l^{\text{word}}\}_{l=1}^L ) \in \mathbb{R}^{K_w \times D}.
\end{equation}

\textbf{Patch Selection Module.}
Given the patch features of the $n$-th frame 
$P_n = \{\phi(p_1^n), \dots, \phi(p_M^n)\}$ with $\phi(p_j^n)\in\mathbb{R}^D$, we first augment each patch with the frame feature $\phi(f^n)$:
\begin{equation}
\tilde{p}_j^n = G_a\big(\mathrm{Concat}(\phi(p_j^n), \phi(f^n))\big),
\end{equation}
then fuse with video-level feature $v$ to produce a saliency score:
\begin{equation}
u_j^n = G_b\big(\mathrm{Concat}(\tilde{p}_j^n, v)\big).
\end{equation}
We select the top-$K_p$ patches per frame based on $u_j^n$, yielding
\begin{equation}
\hat{P}_n = \mathrm{TopK}\big(P_n, \{u_j^n\}_{j=1}^M \big) \in \mathbb{R}^{K_p \times D}.
\end{equation}
Concatenating across all $T$ frames gives
\begin{equation}
\hat{P} = \{\hat{P}_1, \dots, \hat{P}_T\} \in \mathbb{R}^{T \times K_p \times D},
\end{equation}

\textbf{Patch Aggregation Module.}
As shown in Figure \ref{fig:overview2} (b), for each selected word $\hat{w}_l \in \hat{T}$ and selected patch $\hat{p}_j \in \hat{P}$, we compute the similarity:
\begin{equation}
s_{l,j} = \hat{w}_l^\top \hat{p}_j.
\end{equation}
The similarities are normalized with a dynamic temperature $\tau_{\text{dyn}}(\hat{w}_l)$ (as defined previously):
\begin{equation}
a_{l,j} = \frac{\exp(s_{l,j}/\tau_{\text{dyn}}(\hat{w}_l))}{\sum_{k=1}^{L_v} \exp(s_{l,k}/\tau_{\text{dyn}}(\hat{w}_l))}.
\end{equation}
The patch representation guided by word $\hat{w}_l$ is:
\begin{equation}
\hat{v}_{\mathrm{patch}} = \sum_{i=1}^{T}\sum_{l=1}^{L}\sum_{j=1}^{K_p} a_{l,j} \hat{p}_j.
\end{equation}

\begin{table*}[t]
  \caption{Text-to-Video Retrieval Results on DVTMD and capERA(ViT-B/32 and ViT-B/16)}
  \label{tab:t2v}
  \centering
  \small
  \resizebox{\linewidth}{!}{
  \begin{tabular}{l c ccccc c ccccc}
    \toprule
    \multirow{3}{*}{\textbf{Method}} & \multirow{3}{*}{\textbf{Venue} } & \multicolumn{5}{c}{\textbf{ViT-B/32}}  & \multicolumn{5}{c}{\textbf{ViT-B/16}} \\
    \cmidrule(lr){3-7} \cmidrule(lr){8-12}
    & & R@1 $\uparrow$ & R@5 $\uparrow$ & R@10 $\uparrow$ & MdR $\downarrow$ & MnR $\downarrow$ & R@1 $\uparrow$ & R@5 $\uparrow$ & R@10 $\uparrow$ & MdR $\downarrow$ & MnR $\downarrow$ \\
    \midrule
    \multicolumn{12}{l}{\textit{\textbf{Our DVTMD dataset}}} \\
    CLIP \cite{radford2021learning} & ICML’21 & 26.0 & 54.2 & 67.3 & 4.0 & 23.7 & 29.6 & 57.9 & 70.6 & 4.0 & 20.0 \\
    Clip4clip \cite{luo2022clip4clip}& NeuroCom’21 & 41.3 & 72.3 & 82.5 & 2.0 & 9.7 & 44.2 & 75.5 & 84.9 & 2.0 & \textbf{8.7} \\
    Xpool \cite{gorti2022x}& CVPR’22 & 39.0 & 69.6 & 80.5 & 2.0 & 11.8 & - & - & - & - & - \\
    TS2-Net \cite{liu2022ts2} & ECCV’22 & 42.6 & 72.3 & 82.2 & 2.0 & 10.9 & 44.4 & 76.3 & 85.4 & 2.0 & 9.5 \\ 
    DiffusionRet \cite{jin2023diffusionret} & ICCV’23 & 33.3 & 58.5 & 68.2 & 3.0 & 60.4 & 36.7 & 61.8 & 72.3 & 3.0 & 58.4 \\
    UCoFiA \cite{wang2023unified} & ICCV’23 & 42.3 & 70.2 & 81.2 & 2.0 & 11.8 & 46.9 & 74.6 & 85.2 & 2.0 & 11.4 \\
    UATVR \cite{fang2023uatvr} & ICCV’23 & 38.4 & 69.4 & 79.8 & 2.0 & 12.8 & 41.4 & 73.1 & 82.7 & 2.0 & 11.5 \\
    EERCF \cite{tian2024towards} & AAAI'24 & 44.3 & 74.8 & 84.1 & 2.0 & 11.0 & 46.4 & 76.8 & 86.3 & 2.0 & 11.6 \\
    UniAdapter \cite{lu2024uniadapter} & ICLR'24 & 41.1 & 72.8 & 83.0 & 2.0 & \textbf{9.3} & - & - & - & - & - \\
    TempMe \cite{shen2024tempme} & ICLR’25 & 37.5 & 67.5 & 79.5 & 2.0 & 13.9 & 40.4 & 73.2 & 82.7 & 2.0 & 11.9 \\
    % \rowcolor{blue!7} 
    \textbf{TCMA (Ours)} & - & \textbf{45.5} & \textbf{75.3} & \textbf{84.6} & \textbf{2.0} & 10.2 & \textbf{48.5} & \textbf{77.4} & \textbf{86.5} & \textbf{2.0} & 9.4 \\
    \midrule
    \multicolumn{12}{l}{\textit{\textbf{CapERA dataset}}} \\
    CLIP \cite{radford2021learning}& ICML’21 & 8.8 & 24.7 & 36.8 & 19.0 & 52.1 & 8.8 & 26.0 & 38.1 & 18.0 & 48.1 \\
    Clip4clip \cite{luo2022clip4clip}& NeuroCom’21 & 15.0 & 39.5 & 55.7 & 9.0 & 26.9 & \textbf{16.5} & 44.1 & 60.5 & 7.0 & 22.0 \\
    Xpool \cite{gorti2022x}& CVPR’22 & 14.5 & 38.4 & 54.6 & 9.0 & \textbf{24.9} & - & - & - & - & - \\
    TS2-Net \cite{liu2022ts2}& ECCV’22 & 13.8 & 40.5 & 55.3 & 8.0 & 28.0 & 14.7 & 42.2 & 58.7 & 8.0 & 22.7 \\ 
    DiffusionRet \cite{jin2023diffusionret}& ICCV’23 & 11.7 & 32.4 & 45.8 & 13.0 & 63.5 & 13.6 & 35.4 & 48.8 & 11.0 & 54.0 \\
    UCoFiA \cite{wang2023unified}& ICCV’23 & 13.8 & 38.4 & 53.1 & 9.0 & 29.7 & 15.2 & 41.4 & 57.0 & 8.0 & 27.2 \\
    UATVR \cite{fang2023uatvr}& ICCV’23 & 11.9 & 34.2 & 48.6 & 11.0 & 31.8 & 14.1 & 39.3 & 55.9 & 8.0 & 24.6 \\
    EERCF \cite{tian2024towards}& AAAI'24 & 14.8 & 41.6 & 57.7 & 8.0 & 27.9 & \textbf{16.5} & 43.1 & 58.7 & 7.0 & 24.0 \\
    UniAdapter \cite{lu2024uniadapter}& ICLR'24 & \textbf{15.7} & 40.4 & 56.6 & 8.0 & 25.7 & - & - & - & - & - \\
    TempMe \cite{shen2024tempme}& ICLR’25 & 13.4 & 38.2 & 53.8 & 9.0 & 27.5 & \textbf{16.5} & \textbf{44.9} & 59.4 & 7.0 & 21.5 \\
   % \rowcolor{blue!7} 
    \textbf{TCMA (Ours)} & - & 14.9 & \textbf{41.9} & \textbf{57.8} & \textbf{8.0} & 25.6 & 16.1 & 44.6 & \textbf{61.2} & \textbf{7.0} & \textbf{21.2} \\
    \bottomrule
  \end{tabular}
  }
\end{table*}

\subsection{Hierarchical Alignment Loss}
\label{sec:loss}

Contrastive learning is widely adopted in multi-modal learning tasks. Similarly, our framework employs a contrastive objective, treating paired video-text samples as positives and all other non-matching pairs in the batch as negatives.
Given a batch of $B$ video-text pairs $\{(v_i, t_i)\}_{i=1}^B$, we define the contrastive loss in two symmetric directions:
\begin{equation}
\begin{aligned}
L_{v2t}(v,t) &= -\frac{1}{B} \sum_{i=1}^{B} \log \frac{\exp(s(v_i, t_i))}{\sum_{j=1}^{B} \exp(s(v_i, t_j))}, \\ 
L_{t2v}(v,t) &= -\frac{1}{B} \sum_{i=1}^{B} \log \frac{\exp(s(v_i, t_i))}{\sum_{j=1}^{B} \exp(s(v_j, t_i))},
\end{aligned}
\end{equation}
where $v\in \mathbb{R}^{B\times D}$ and $t\in \mathbb{R}^{B\times D}$ are the video and text feature matrices, and $s(\cdot, \cdot)$ denotes cosine similarity.  
The $v2t$ term encourages each video to be most similar to its corresponding text, while $t2v$ ensures each text is most similar to its corresponding video.  
By jointly optimizing both directions, the model learns robust cross-modal alignments.

In addition to the contrastive objective, we incorporate a Pearson correlation constraint to regularize feature structures.  
This term encourages same-channel consistency, where the $d$-th channel of video features aligns with the $d$-th channel of text features, while reducing cross-channel redundancy. The Pearson correlation coefficient between two vectors $x$ and $y$ is defined as
\[
\rho_p(x,y) = \frac{(x-\bar{x})^\top(y-\bar{y})}{\|x-\bar{x}\|\|y-\bar{y}\|},
\]
and the corresponding Pearson distance is
\[
d_p(x,y) = 1 - \rho_p(x,y).
\]
The Pearson regularization term over a batch of features is then defined as
\begin{align}
L_{Pearson}(v,t) 
&= \beta \sum_{d=1}^{D} \big\| d_p(v_{:,d}, t_{:,d}) \big\|^2  \notag \\
&\quad + \alpha \sum_{d_1=1}^{D} \sum_{d_2 \ne d_1} 
   \big\| \rho_p(v_{:,d_1}, t_{:,d_2}) \big\|^2,
\end{align}
where $||\cdot||$ denotes the L2-norm and $\beta$ and $\alpha$ control the strength of regularization.  
The first term enforces same-channel alignment, while the second discourages cross-channel redundancy.  
Thus, we define a general loss function that combines both contrastive and Pearson terms:
\begin{equation}
L(v,t) = L_{v2t}(v,t) + L_{t2v}(v,t) + L_{Pearson}(v,t).
\end{equation}

\textbf{Video-Level Alignment.}  
The video-level loss captures global semantics across all frames:
\begin{equation}
    L_{\text{video}} = L(\hat{v}_{\text{video}}, \phi(t)),
\end{equation}
where $\hat{v}_{\text{video}}$ is the mean-pooled video representation and $\phi(t)$ is the text embedding.

\textbf{Frame-Level Alignment.}  
The frame-level loss models the relevance of individual frames guided by sentence-level features:
\begin{equation}
    L_{\text{frame}} = L(\hat{v}_{\text{frame}}, \phi(t)),
\end{equation}
where $\hat{v}_{\text{frame}}$ is obtained from sentence-guided frame aggregation.

\textbf{Patch-Level Alignment.}  
The patch-level loss captures fine-grained correspondence between salient patches and words:
\begin{equation}
    L_{\text{patch}} = L(\hat{v}_{\text{patch}}, \phi(t)),
\end{equation}
where $\hat{v}_{\text{patch}}$ is the output of word-guided patch aggregation.

\textbf{Overall Hierarchical Objective.}  
Finally, the overall loss combines all levels in a weighted manner:
\begin{equation}
\mathcal{L}_{\text{hier}} = 
\lambda_{\text{video}} \, L_{\text{video}}
+ \lambda_{\text{frame}} \, L_{\text{frame}}
+ \lambda_{\text{patch}} \, L_{\text{patch}},
\end{equation}
where $\lambda_{\text{video}}$, $\lambda_{\text{frame}}$, and $\lambda_{\text{patch}}$ are hyperparameters controlling the contribution of each hierarchical level.

\section{Experiments}
In this section, we conduct extensive experiments to evaluate the effectiveness of our TCMA model. We introduce our experimental settings in Section \ref{sec:setting}, compare it with state-of-the-art methods in Section \ref{sec:compare}, perform ablation studies to analyze the contribution of each component in Section \ref{sec:ablation}, and provide qualitative results in Section \ref{sec:qual}.
% to demonstrate its capability in capturing fine-grained and contextual information in UAV video scenarios in Section \ref{sec:qual}.

\begin{table*}[t]
  \caption{Video-to-Text Retrieval Results on DVTMD and capERA (ViT-B/32 and ViT-B/16)}
  \label{tab:v2t}
  \centering
  \small
\resizebox{0.95\linewidth}{!}{
  \begin{tabular}{l c ccccc ccccc}
    \toprule
    \multirow{3}{*}{\textbf{Method}} & \multirow{3}{*}{\textbf{Venue}} & \multicolumn{5}{c}{\textbf{ViT-B/32}}  & \multicolumn{5}{c}{\textbf{ViT-B/16}} \\
    \cmidrule(lr){3-7} \cmidrule(lr){8-12}
    & & R@1 $\uparrow$ & R@5 $\uparrow$ & R@10 $\uparrow$ & MdR $\downarrow$ & MnR $\downarrow$ & R@1 $\uparrow$ & R@5 $\uparrow$ & R@10 $\uparrow$ & MdR $\downarrow$ & MnR $\downarrow$ \\
    \midrule
    \multicolumn{12}{l}{\textit{\textbf{Our DVTMD dataset}}} \\
    CLIP \cite{radford2021learning}& ICML'21 & 28.5 & 60.4 & 74.9 & 4.0 & 10.0 
                   & 30.8 & 64.3 & 78.2 & 3.0 & 8.3 \\
    TS2-Net \cite{liu2022ts2}& ECCV'22 & 38.5 & 69.7 & 81.2 & 2.0 & 8.4 
                       & 41.0 & 72.4 & 84.9 & 2.0 & 6.8 \\
    UCoFiA \cite{wang2023unified}& ICCV'23 & 37.2 & 68.7 & 80.9 & 2.0 & 9.8 
                      & 40.5 & 72.9 & 83.8 & 2.0 & 8.1 \\
    UATVR \cite{fang2023uatvr}& ICCV'23 & 34.2 & 66.4 & 79.0 & 3.0 & 9.6 
                      & 36.7 & 70.1 & 81.4 & 2.0 & 7.8 \\
    EERCF \cite{tian2024towards}& AAAI'24 & 40.0 & 69.9 & 82.6 & 2.0 & 8.4 
                      & 41.8 & 73.8 & 84.4 & 2.0 & 7.1 \\
    UniAdapter \cite{lu2024uniadapter}& ICLR'24 & 41.0 & \textbf{71.7} & 82.9 & 2.0 & \textbf{7.0} 
                          & - & - & / & - & - \\
    TempMe \cite{shen2024tempme}& ICLR'25 & 37.1 & 66.3 & 77.9 & 3.0 & 9.8 
                       & 38.7 & 68.3 & 81.3 & 2.0 & 8.2 \\
   % \rowcolor{blue!7}     
    \textbf{TCMA (Ours)} & -& \textbf{42.8} & 71.2 & \textbf{83.4} & \textbf{2.0} & 8.0  & \textbf{45.9} & \textbf{74.3} & \textbf{85.3} & \textbf{2.0} & \textbf{6.4} \\
    \midrule
    \multicolumn{12}{l}{\textit{\textbf{CapERA dataset}}} \\
     CLIP \cite{radford2021learning} & ICML'21 & 4.6 & 18.6 & 32.0 & 21.0 & 43.8 & 4.5 & 18.8 & 32.6 & 21.0 & 41.6 \\
    TS2-Net \cite{liu2022ts2}& ECCV'22 & 9.7 & 31.0 & 46.4 & 12.0 & 29.0 & 10.8 & 35.3 & 51.4 & 10.0 & 24.6 \\
    UCoFiA \cite{wang2023unified}& ICCV'23 & 8.5 & 29.0 & 42.3 & 14.0 & 33.3 & 9.9 & 32.1 & 47.2 & 12.0 & 28.7 \\
    UATVR \cite{fang2023uatvr}& ICCV'23 & 8.2 & 27.3 & 42.4 & 15.0 & 35.2 & 9.9 & 32.4 & 48.7 & 11.0 & 25.4 \\
    EERCF \cite{tian2024towards}& AAAI'24 & 10.5 & 32.2 & 47.0 & 12.0 & 29.6 & 10.7 & 36.1 & 50.8 & 10.0 & 25.3 \\
    UniAdapter \cite{lu2024uniadapter}& ICLR'24 & 11.0 & 34.2 & 48.7 & 11.0 & \textbf{26.6} & - & - & - & - & - \\
    TempMe \cite{shen2024tempme}& ICLR'25 & 11.0 & 32.2 & 46.3 & 12.0 & 28.1 & 11.9 & 36.4 & 52.2 & 10.0 & \textbf{22.3} \\
    % \rowcolor{blue!7} 
    \textbf{TCMA (Ours)} & -& \textbf{12.7} & \textbf{34.6} & \textbf{49.6} & \textbf{11.0} & 31.9 & \textbf{15.0} & \textbf{39.3} & \textbf{53.5} & \textbf{9.0} & 26.2 \\
    \bottomrule
  \end{tabular}
  }
\end{table*}

\subsection{Experimental Settings}
\label{sec:setting}
\textbf{Evaluation Metrics.} We evaluate our TCMA model on standard text-video retrieval benchmarks using Recall at rank $K$ (R@1, R@5, R@10), Mean Rank (MnR), and Median Recall (MdR).

\textbf{Implementation Details.}
All experiments are conducted on a single H20-NVLink GPU (96GB) and implemented in PyTorch. We initialize our TCMA model with CLIP \cite{radford2021learning} pre-trained weights. For the visual encoder, we adopt CLIP’s ViT-B/32 and ViT-B/16 backbones, where the feature dimension $D$ for both visual and textual embeddings is set to 512. For optimization, we use Adam with a cosine warm-up schedule. The learning rates are set to $1\times10^{-7}$ for the pre-trained visual and text encoders, and $1\times10^{-4}$ for the newly added modules. The batch size is fixed at 64.

Regarding the method-specific parameters, in the patch selection module, we retain the top-$K_p = 3$ most salient patches per frame, and in the word selection module, the top-$K_w = 8$ most salient words are chosen. We uniformly sample $T = 12$ frames per video, and the maximum text length is capped at 32 tokens. Unless otherwise specified, the hyperparameters in our loss functions are empirically set as follows: the intra-modal loss weight is fixed at $\alpha = 0.05$, and the total loss incorporates $\beta = 0.001$ with a weighting ratio of $\lambda_{V_{video}} : \lambda_{v_{frame}} : \lambda_{v_{patch}} = 5 : 5 : 1$.

\begin{table}[t]
\caption{Ablation on Hierarchical Levels.}
\label{tab:ablation_hier}
\centering
\small
\resizebox{ \linewidth}{!}{ 
\begin{tabular}{lcccccc}
\toprule
\multirow{2}{*}{Setting} & \multicolumn{3}{c}{Text-to-Video} & \multicolumn{3}{c}{Video-to-Text}  \\
\cmidrule(lr){2-4} \cmidrule(lr){5-7}
 & R@1 & R@5 & R@10 & R@1 & R@5 & R@10 \\
\midrule
Video only & 43.9 & 75.1 & 83.8 & 41.0 & 70.2 & 82.2 \\
Video+Frame & 45.5 & 75.3 & 84.6 & 42.5 & 71.2 & 83.4 \\
Video+Patch & 43.6 & 74.8 & 84.2 & 41.3 & 70.0 & 82.2 \\
Video+Frame+Patch & \textbf{45.5} & \textbf{75.3} & \textbf{84.6} & \textbf{42.8} & \textbf{71.2} & \textbf{83.4} \\
\bottomrule
\end{tabular}%
}
\end{table}%

\begin{table}[t]
\caption{Ablation study on word selection ($K_w$).}
\label{tab:word_selection_ablation}
\centering
\small
\resizebox{ \linewidth}{!}{ 
\begin{tabular}{lcccccc}
\toprule
\multirow{2}{*}{Setting} & \multicolumn{3}{c}{Text-to-Video} & \multicolumn{3}{c}{Video-to-Text}  \\
\cmidrule(lr){2-4} \cmidrule(lr){5-7}
 & R@1 & R@5 & R@10 & R@1 & R@5 & R@10 \\
\midrule
$K_w=4$   & 45.4 & 75.3 & 84.6 & 42.3 & 71.1 & 83.3 \\
$K_w=8$   & \textbf{45.5} & \textbf{75.3} & \textbf{84.6} & \textbf{42.8} & \textbf{71.2} & \textbf{83.4} \\
$K_w=16$  & 45.4 & 75.2 & 84.5 & 42.8 & 71.2 & 83.4 \\
w/o word sel. & 44.6 & 74.7 & 83.9 & 41.3 & 70.7 & 82.5 \\
\bottomrule
\end{tabular}%
}
\end{table}%

\subsection{Comparisons With State-of-the-Art Methods}
\label{sec:compare}
We compare TCMA against a range of state-of-the-art text–video retrieval methods, including global-matching approaches \cite{radford2021learning, luo2022clip4clip}, fine-grained alignment frameworks \cite{liu2022ts2}, generative models \cite{jin2023diffusionret}, parameter-efficient approaches \cite{lu2024uniadapter}, and coarse-to-fine alignment models for natural video understanding \cite{wang2023unified, tian2024towards}.

Global-matching methods such as CLIP \cite{radford2021learning} and Clip4Clip \cite{luo2022clip4clip} achieve strong performance on generic video benchmarks, but they underperform on UAV datasets where targets are small, backgrounds are uniform, and global video representations lack sufficient discriminative capacity. Fine-grained methods such as TS2-Net \cite{liu2022ts2} leverage region-level cues but fail to capture holistic global context, which is critical for integrating information in UAV scenes. Generative approaches like DiffusionRet \cite{jin2023diffusionret} incur substantial computational overhead and exhibit unstable retrieval quality, as evidenced by their significantly degraded MnR on DVTMD. Finally, recent coarse-to-fine models, including UCoFiA \cite{wang2023unified} and EERCF \cite{tian2024towards}, improve cross-modal alignment but are primarily optimized for natural video domains. They struggle to adapt to aerial scenarios, where objects vary greatly in scale, noise is abundant, and inter-class distinctions are subtle.

In contrast, TCMA is purpose-built for UAV video retrieval. Its hierarchical aggregation fuses both global and local information, token selection modules filter out irrelevant noise in cluttered aerial scenes, and the dynamic temperature mechanism adapts to variations in object and scene scale. Together, these designs directly tackle the challenges of small-object resolution, fine-grained motion modeling, and context disambiguation, yielding robust performance in aerial video retrieval.

Tables \ref{tab:t2v} and \ref{tab:v2t} show that TCMA achieves consistent improvements over all baselines on both DVTMD and capERA datasets. With the ViT-B/32 backbone, TCMA achieves 45.5\% R@1 and 75.3\% R@5 in text-to-video retrieval, improving by +1.2\% and +0.5\% over the previous best method EERCF \cite{tian2024towards}. Using the stronger ViT-B/16 backbone, the margin further widens as TCMA reaches 48.5\% R@1. A similar trend holds in video-to-text retrieval. Our TCMA with ViT-B/16 obtains 45.9\% R@1, surpassing EERCF by +4.1\%, the largest improvement among all settings. The retrieval on capERA is more challenging due to the weak text-video correspondence as mentioned before. However, TCMA maintains the best performance among all methods, raising R@10 by +1.1\% on ViT-B/16 and reducing MnR to 21.2.

\subsection{Ablation Studies}
\label{sec:ablation}

To better understand the contributions of each component in our TCMA framework, we conduct ablation studies on the DVTMD dataset with ViT/32 backbone.

\textbf{Hierarchical Aggregation.} We first investigate the effect of progressively introducing different levels of alignment. Starting from a video-only baseline that uses mean-pooled representations, we add frame-level alignment via sentence-guided aggregation, then further incorporate patch-level alignment guided by word features. The results in Table~\ref{tab:ablation_hier} show that each additional level consistently improves retrieval performance. The full TCMA model, which combines all three levels, achieves the best overall performance.

\begin{table}[t]
\caption{Ablation on patch selection ($K_p$).}
\label{tab:k_p_ablation}
\centering
\small
\resizebox{ \linewidth}{!}{ 
\begin{tabular}{lcccccc}
\toprule
\multirow{2}{*}{Setting} & \multicolumn{3}{c}{Text-to-Video} & \multicolumn{3}{c}{Video-to-Text} \\
\cmidrule(lr){2-4} \cmidrule(lr){5-7}
 & R@1 $\uparrow$ & R@5 $\uparrow$ & R@10 $\uparrow$ & R@1 $\uparrow$ & R@5 $\uparrow$ & R@10 $\uparrow$ \\
\midrule
$K_p=2$         & 45.6 & 75.3 & 84.0 & 42.7 & 71.7 & 83.1 \\
$K_p=3$         & 45.5 & \textbf{75.3} & \textbf{84.6} & \textbf{42.8} & 71.2 & 83.4 \\
$K_p=4$         & \textbf{45.9} & 75.3 & 83.9 & 42.7 & 71.5 & 83.2 \\
$K_p=6$         & 45.7 & 74.7 & 83.9 & 42.7 & 71.6 & 83.0 \\
$K_p=8$         & 45.7 & 74.8 & 83.8 & 42.6 & 71.7 & 83.2 \\
w/o patch sel.  & 45.1 & 74.8 & 84.1 & 42.7 & \textbf{71.8} & \textbf{83.7} \\
\bottomrule
\end{tabular}
}
\end{table}

\begin{table}[t]
\caption{Ablation on Dynamic Temperature $\tau_{\text{dyn}}$.}
\label{tab:ablation_temp}
\centering
\small
\resizebox{ \linewidth}{!}{ 
\begin{tabular}{lcccccccccc}
\toprule
\multirow{2}{*}{Setting} & \multicolumn{3}{c}{Text-to-Video} & \multicolumn{3}{c}{Video-to-Text} \\
\cmidrule(lr){2-4} \cmidrule(lr){5-7}
 & R@1 $\uparrow$ & R@5 $\uparrow$ & R@10 $\uparrow$ & R@1 $\uparrow$ & R@5 $\uparrow$ & R@10 $\uparrow$ \\
\midrule
w/o $\tau_{\text{dyn}}$ & \textbf{45.7} & 74.8 & 84.0 & 42.0 & 71.2 & 82.8 \\
with $\tau_{\text{dyn}}$ & {45.5} & \textbf{75.3} & \textbf{84.6} & \textbf{42.8} & \textbf{71.2} & \textbf{83.4} \\
\bottomrule
\end{tabular}
}
\end{table}

\begin{figure*}[!t]
\centering
\includegraphics[width=0.8\linewidth]{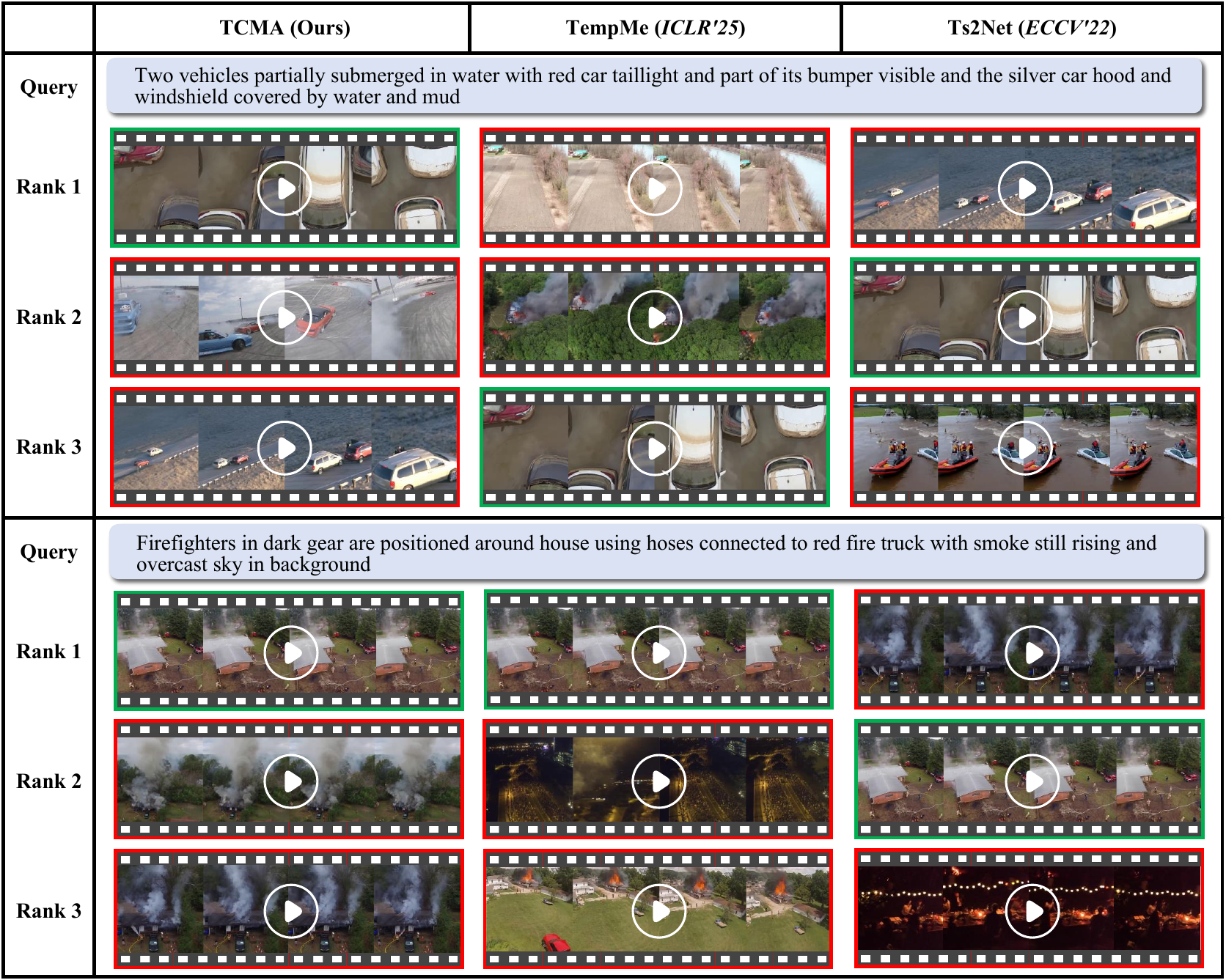}
\caption{Visualization of text–video retrieval examples. Our retrieval results are compared with other state-of-the-art methods and ranked according to similarity scores. Green boxes indicate the ground truth, while red boxes denote incorrect results.}
\label{fig:qualitative}
\end{figure*}

\textbf{Word and Patch Selection.} We then analyze the role of word and patch selection. Without any filtering, all words are aligned with all patches, which introduces noise from uninformative tokens. Selecting only the most salient patches already improves performance, and word selection brings similar benefits. The results in Table~\ref{tab:k_p_ablation} and Table~\ref{tab:word_selection_ablation} show When both strategies are applied jointly, as in our final design, the model achieves the best retrieval accuracy.

\textbf{Dynamic Temperature.} Finally, we evaluate the impact of the dynamic temperature coefficient $\tau_{\text{dyn}}$ and record the results in Table~\ref{tab:ablation_temp}. Using a fixed temperature produces competitive results, but adapting the temperature dynamically yields consistent improvements across recall metrics.

\subsection{Qualitative Results}
\label{sec:qual}
We visualize some retrieval examples on our DVTMD set for text-to-video retrieval in Figure \ref{fig:qualitative}. In the first case, our model accurately identifies the number of objects (“two vehicles”) and their respective attributes (“red” and “silver”), while also distinguishing “mud” from regular road surfaces. In the last case, the model successfully retrieves a video containing both human figures (“firefighters in dark gear”) and relevant objects (“red fire truck”), highlighting its capability to jointly reason over actions and contextual elements.

\section{Conclusion}
In this paper, we propose a drone text-video retrieval framework that jointly captures global semantics and fine-grained details through multi-granularity alignment. We mainly establish the TCMA model, which integrates global video-sentence alignment, sentence-guided frame aggregation, and word-guided patch alignment, and contribute the fine-grained DVTMD dataset. The results show that for UAV videos with large spatial scales, diverse objects, and complex backgrounds, retrieval accuracy can be significantly improved by using Top-K Word and Patch selection to filter irrelevant content and a dynamic temperature module to adapt attention sharpness. The qualitative and quantitative results on DVTMD and CapERA indicate that multi-granularity alignment with adaptive attention may be a promising direction for advancing drone text-video retrieval.

\bibliographystyle{IEEEtran}
\bibliography{references}


%\vfill


\end{document}